\documentclass[letterpaper, 10 pt, conference]{ieeeconf}  
\usepackage{hyperref}  

\IEEEoverridecommandlockouts                              

\usepackage{graphics} 
\usepackage{epsfig} 
\usepackage{mathptmx} 
\usepackage{times} 
\usepackage{amsmath} 
\usepackage{amssymb}  
\usepackage{clrscode}
\usepackage{algorithmic}
\usepackage{subfig}
\usepackage{float}
\usepackage{array}
\usepackage{multirow}
\newcolumntype{L}[1]{>{\raggedright\let\newline\\\arraybackslash\hspace{0pt}}m{#1}}
\newcolumntype{C}[1]{>{\centering\let\newline\\\arraybackslash\hspace{0pt}}m{#1}}
\newcolumntype{R}[1]{>{\raggedleft\let\newline\\\arraybackslash\hspace{0pt}}m{#1}}

\title{\LARGE \bf
Baidu Apollo Auto-Calibration System - An Industry-Level Data-Driven and Learning based Vehicle Longitude Dynamic Calibrating Algorithm
}

\author{Fan Zhu$^{1, \dag}$, Lin Ma$^{2, \dag}$, Xin Xu$^{2, \dag}$, Dingfeng Guo$^{2, \dag}$, Xiao Cui$^{2, \dag}$, and Qi Kong$^{1, *}$
\thanks{\dag Authors that contribute equally in this manuscript.}
\thanks{* Author that corresponds in this manuscript.}
\thanks{$^{1}$Baidu USA LLC, Sunnyvale, CA 94089, USA}
\thanks{$^{2}$Baidu Inc., Haidian District, Beijing, China}
\thanks{
Fan Zhu, fanzhu@baidu.com;
Lin Ma, malin08@baidu.com;
Xin Xu, xuxin11@baidu.com;
Dingfeng Guo, guodingfeng@baidu.com;
Xiao Cui, cuixiao@baidu.com;
Qi Kong, kongqi02@baidu.com}
}

\begin{document}

\maketitle
\thispagestyle{empty}
\pagestyle{empty}

\begin{abstract}

For any autonomous driving vehicle, 
control module determines its road performance and safety, 
i.e. its precision and stability should stay within a carefully-designed range. 
Nonetheless, control algorithms require vehicle dynamics (such as longitudinal dynamics) as inputs, which, unfortunately, are obscure to calibrate in real time. 
As a result, to achieve reasonable performance, most, if not all, research-oriented autonomous vehicles do manual calibrations in a one-by-one fashion. 
Since manual calibration is not sustainable once entering into mass production stage for industrial purposes, we here introduce a machine-learning based auto-calibration system for autonomous driving vehicles.

In this paper, we will show how we build a data-driven longitudinal calibration procedure using machine learning techniques. 
We first generated offline calibration tables from human driving data. 
The offline table serves as an initial guess for later uses and it only needs twenty-minutes data collection and process. 
We then used an online-learning algorithm to appropriately update the initial table (the offline table) based on real-time performance analysis. 

This longitudinal auto-calibration system has been deployed to more than one hundred Baidu Apollo self-driving vehicles (including hybrid family vehicles and electronic delivery-only vehicles) since April 2018. 
By August 27, 2018, it had been tested for more than two thousands hours, ten thousands kilometers (6,213 miles) and yet proven to be effective.

\end{abstract}

\section{INTRODUCTION}
Autonomous driving technology has been a particular interest for both industrial and research communities in the last a few years.
Baidu, one of the leading companies in this field, has been putting great efforts into building an open community with its open-source self-driving solution since 2017 \cite{Apollo2017, Fan2018, Fan2018_2}. 
With years' development (Baidu began its autonomous driving research since 2013), 
we had tested hundreds of vehicles with tens of thousand hours and with multiple generations of algorithms. 
As the number of deployed vehicles keeps increasing, we soon found that manually calibrating (we only refer to longitudinal vehicle dynamics calibration in this paper) each vehicle is infeasible. Manual calibration involves considerable labours, which, in other words, predicts large amount of time and great potential for man-made mistakes. Further, vehicle dynamics usually vary noticeably during driving (i.e. loads changes, vehicle parts worn out over time, surface friction \cite{Canudas-de-Wit2003}), and manual calibration cannot possibly cover them. Taken together, it is surprising that most research papers did not attempt to address this manual-calibration puzzle. One could attribute this to the fact that most research-oriented projects only focus on very few vehicles. To solve this industrial-specific problem, in this paper we propose a novel auto-calibration (for longitudinal dynamics) system based on both offline  model and online learning.

Previous researches have investigated longitudinal control algorithms extensively, and regarded it as a challenging problem \cite{Khodayari2010,Rajamani2011,Pack2011}. One major challenge lies in a mission impossible --- to establish an accurate longitudinal vehicle dynamics in real time \cite{Chen2011,Falcone2007,Martinez2007}. A control model with accurate longitudinal vehicle dynamics bridges gap between desired speeds and vehicle throttling/braking commands. A popular solution is to establish this state-space relation based on Newton formula \cite{Bageshwar2004,Levinson2011,Raffo2009, Attia2014}. However, transmission system, power-train system, and actuator system are all very complex. It is not only difficult to model all these systems but also it requires unacceptable computing time as more systems involved.

In this paper, we address this challenging topic from another perspective. We consider the entire longitudinal control algorithm as an end-to-end problem, which can be solved in two steps:
(1) Based on human driving data, we first generate a calibration table, which takes throttle, brake, and speed as inputs, and outputs acceleration. 
(2) Then a sophisticated on-line algorithm updates the table to properly cover the varying vehicle dynamics. 
Our results show that the auto-calibration system does save considerable amount of time and improves control accuracy. 
The calibration is also automated and intelligent, due to which, is suitable for mass-scale self-driving vehicle deployment.

\section{METHOD}\label{sec:method}

\subsection{Apollo Autonomous Driving Control Module}
\begin{figure}[thpb]
	\small
	\centering
	\includegraphics[width = 1.0\linewidth]{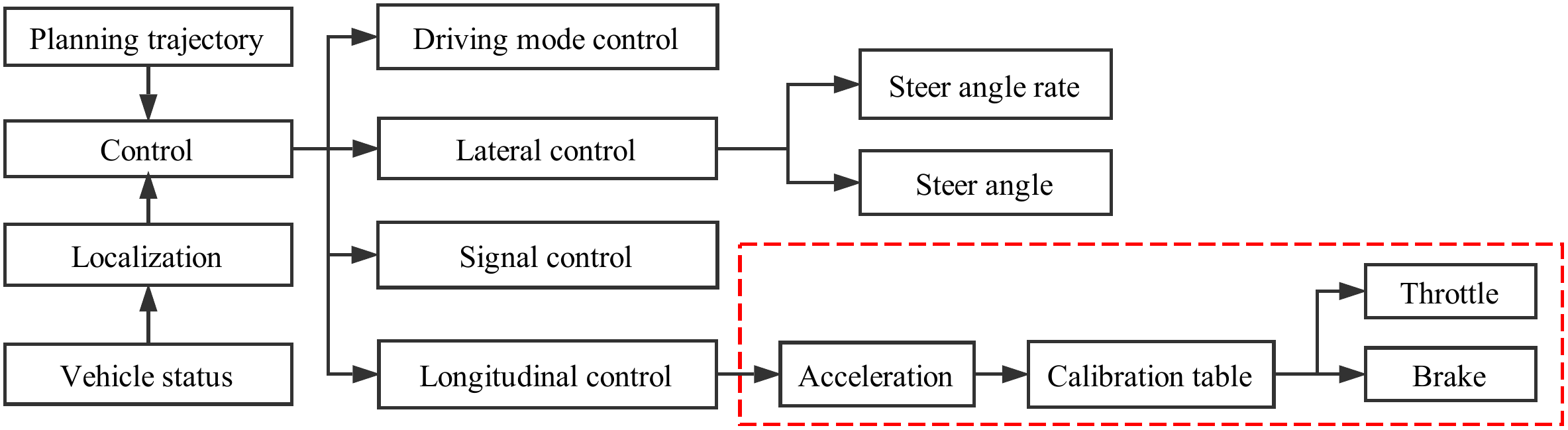}
	\caption{Apollo control framework}
	\label{controlframework}
\end{figure}

Before launching into the methodological detail of this paper,
it is useful to briefly review the current architecture of Apollo control module.
Apollo control module is designed to track trajectories generated by Apollo planning module, and minimize tracking errors. It takes trajectory, vehicle position, and vehicle status as inputs, and sends steering, braking, and throttling commands to manipulate vehicle.
Apollo control module includes lateral algorithm (which generates steering commands) and longitudinal algorithm (which generates braking/throttling commands), see Fig.\ref{controlframework} for details.
The longitudinal calibration algorithm, the topic of this paper, calculates throttle/brake values in the form of Eq. \ref{eq:calitable}. Note that acceleration ($acc$) and speed ($v$) are known variables given by longitudinal algorithm. This table hence simply maps ($acc, v$) to throttling/braking commands ($cmd$).
An offline auto-calibration algorithm is used to generate an initial table for each vehicle. 
To cover varying vehicle dynamics, Apollo control module compensates offline table with an online learning algorithm. 
Together, the offline model and online algorithm guarantee the stability and accuracy of Apollo control module.

\begin{equation}
T:{acc, v}\to cmd\label{eq:calitable}
\end{equation}
where $T$ refers to calibration table, $acc$ is desired acceleration, $v$ is current vehicle speed, and $cmd$ is control command, i.e. throttling/braking value.

\subsubsection{Offline Model}
Offline model generates an initial calibration table from manual driving data that best reflects vehicle longitudinal performance at the time of driving.
The workflow of offline algorithm includes:
(1) collecting human driving data,
(2) preprocessing the data and select input features,
(3) generating calibration table through machine learning models.

\subsubsection{Online Learning}
Online algorithm instantly adjusts the offline table based on real-time feedback in self-driving mode. 
It aims to best match current vehicle dynamics based on offline model established from manual driving data.
Several challenges this algorithm has to handle includes frequent vehicle load changes and mechanical components fatigue in the long run.
The workflow of online algorithm includes: 
(1) collecting vehicle status and feedback in real time,
(2) preprocessing and filter data, 
(3) adjusting calibration table accordingly.

\subsection{Data Preprocessing}\label{sec:datapreprocessing}
Since real-world data always comes with noises, 
it is essential to preprocess data for both 
offline model and online learning algorithm. 
Note that due to different data sources, 
separate data preprocessing methods are used 
for offline model and online learning algorithm.

\subsubsection{Offline Model Data Preprocessing}\label{sec:offmodel}
Offline model takes human driving data as input. 
Throttling/braking values, speed, and longitudinal acceleration are chosen as features for this model. 
Note that throttling model and braking model are trained separately.

Since human driving data are naturally noisy and non-uniform (i.e. speed/acceleration does not distribute evenly during driving), 
Apollo control module takes a few steps to clean the data. First, a mean filter in Eq. \ref{eq:filter1} is used to smooth data. 
Further, Eq. \ref{eq:off_theta} is applied to remove data with lateral swing. Last but not least, Eq. \ref{eq:filter3} is enforced to remove outliers.

\begin{equation}
y = \cfrac{(x_{t-1} + x_{t-2} +\cdots+ x_{t-N})}{N}
\label{eq:filter1}
\end{equation}
where $N$ is the mean filter window size, $x_{t-1}$, $x_{t-2}$, $...$, $x_{t-N}$ are data from time $t-1$ to $t-N$.

\begin{equation}
\left| \theta \right| < \delta_{steer} \label{eq:off_theta}
\end{equation}
where $\theta$ refers to steering wheel angle, $\delta_{steer}$
refers to steering wheel angle threshold.

\begin{equation}
\cfrac{\left| x - x_{mean}\right|} {x_{std}} > 1 \to outliers\label{eq:filter3}
\end{equation}
where $x$ is data to be processed, $x_{mean}$ is mean of the data, $x_{std}$ is standard deviation of the data.

\subsubsection{Online Learning Data Preprocessing}
Online learning algorithm frequently adjusts calibration table built by 
offline model, on the basis of real-time vehicular status 
feedback in self-driving mode. 
The selected features for this algorithm are: 
a) throttling/braking commands; b) speed; 
c) desired acceleration; d) actual acceleration. 

Again, only data complying with Eq. \ref{eq:off_theta} are used.
Meanwhile, since certain differential techniques are involved
 (see next Section for details), we use Eq. \ref{eq:mono} to guarantee
  monotonicity throughout sequences of control commands.
\begin{equation}
\cfrac{T'[cmd_{i}][v_{x}] - T'[cmd_{j}][v_{x}]} {cmd_{i} - cmd_{j}} < 0 \text{, } \forall cmd_{i}, cmd_{j}, v_x \label{eq:mono}
\end{equation}
where $T'$ refers to updated calibration table; $T'[cmd_{i}][v_{x}]$ indexes the acceleration corresponding to control command $cmd_{i}$ and speed $v_{x}$.

As to real-time data acquisition, one would expect a response delay
(i.e. the time between throttling/braking commands being sent
and corresponding acceleration being executed) caused by vehicular actuator. 
Such delay inevitably affects the quality of data, a 
common and useful technique is to collect data from sensors 
with a reasonable delay estimation \cite{Sukkarieh1999}. 
In our case, we take acceleration measured by IMU (Inertial Measurement Unit), 
200 milliseconds after sending throttling/braking command, 
as the actual acceleration driven by that command. A Butterworth low pass filter, with an order of 3 and 
cutoff frequency of 2 Hz, is then used to remove high-frequency fluctuations. 
In case that 200 milliseconds' delay estimation may not always be
appropriate, Eq. \ref{eq:history_consis} is used to further ensure data consistency.
\begin{equation}
\left| cmd_{ref} - cmd_{ref+\Delta_{t}} \right| < \delta_{cmd\_gap} \text{, } \Delta_{t} = -100ms, \cdots ,100 ms\label{eq:history_consis}
\end{equation}
where $cmd_{ref}$ is the control command to be updated, i.e. command applied 200 milliseconds ago;
$cmd_{ref+\Delta_{t}}$ is control command at time $ref+\Delta_{t}$;
$\delta_{cmd\_gap}$ is the maximal command perturbations allowed.
Note that $cmd_{ref}$ is to be processed only if every element within its neighbourhood [$cmd_{ref - 100ms}$, $cmd_{ref + 100ms}$] satisfies Eq. \ref{eq:history_consis}. 
Finally, Eq. \ref{eq:acc_consis} is enforced on sensor data ($v_{k}$, $a_{k}$) to remove senseless noises.
\begin{equation}
(v_{ref}-v_{k})\times(a_{ref}-a_{k}) > 0\label{eq:acc_consis}
\end{equation}

\subsection{Offline Model Algorithm} \label{sec:offline_algorithm}

\begin{table}
\begin{center}
\rule{\linewidth}{1pt}
\small{
\begin{codebox} 
\Procname{$\proc{Pseudo Algorithm 1: Offline auto-calibration}$}
\zi \textbf{Input:} $cmd_{}$: control command (throttle/brake)
\zi \textbf{Input:} $speed_{}$: current speed
\zi \textbf{Input:} $acc_{}$: acceleration from IMU
\zi \textbf{Input:} $\theta$: steering wheel angle
\\
\li \If $\left| \theta_{} \right| > \delta_{steer}$ \label{li:pa1_start}
\li \Then remove current sample  \label{li:pa1_filter}
    \End
\li \For each $sample$
\li 	      \Then determine [v][cmd] grid based on sample$[v_{i}][cmd_{j}]$
\li                      store $sample_{acc}$ to corresponding [v][cmd] grid 
    	      \End
    \End \label{li:pa1_uniform2}
\li  \For each [v][cmd] grid
\li \Then uniform sample numbers in each [v][cmd] grid
\li           remove outlier $sample_{acc\_outlier}$ from each grid \label{li:pa1_filter2}
    \End \label{li:pa1_filter3}
\li Throttle Model $\gets$ $cmd_{throttle}, acc_{filtered}, v_{filtered}$ \label{li:pa1_model1}
\li Brake Model $\gets$ $cmd_{brake}, acc_{filtered}, v_{filtered}$ \label{li:pa1_model2}
\end{codebox}
}
\rule{\linewidth}{1pt}
\end{center}
\end{table}

Pseudo Algorithm 1 illustrates the procedure of offline model training process.
Data is first preprocessed as described in Section \ref{sec:offmodel} 
(Line \ref{li:pa1_start} to Line \ref{li:pa1_filter3}), after which, 
different models for throttling and braking are constructed
 (Line \ref{li:pa1_model1} to Line \ref{li:pa1_model2}).
A standard three-layer feedforward neural network is used in offline algorithm model, 
with a sigmoid function as the activation function and mean square error as the cost function.
The network is built using Tensorflow with a build-in Adam optimizer.
In comparison, we have evaluated multiple traditional machine learning regression methods for offline model:
\begin{itemize} 

\item \textit{Gaussian Process Regression} \cite{Rasmussen2004}, with polynomial kernel. It is a model-free algorithm used to solve regression problems in various domains \cite{Zhu2012, Zhu2014, Sieberts2016, Zhu2018}.

\item \textit{Linear Regression}, a basic regression model with linear kernel.

\item \textit{Support Vector Machine} \cite{Hearst1998}, with polynomial kernel.

\item \textit{M5P} \cite{Quinlan1992}, a M5 model trees.

\item \textit{Random Forest} \cite{Breiman2001}, with one hundred iterations.

\end{itemize}

These methods were implemented and evaluated with WEKA \cite{Hall2009}. Detailed settings and parameters can be found in WEKA.


\subsection{Online Learning Algorithm}\label{sec:onlinelearningalg}
For each control cycle, online learning algorithm updates calibration table based on real-time vehicular feedback. Once adjusted, new table takes effect immediately from the next control cycle. Pseudo Algorithm 2 gives the  implementation detalis of real-time calibration process.

\begin{table}
\begin{center}
\rule{\linewidth}{1pt}
\small{
\begin{codebox} 
\Procname{$\proc{Pseudo Algorithm 2: Online auto-calibration }$}
\zi \textbf{Input:} $cmd_{ref, k}$: control command at time $k$
\zi \textbf{Input:} $v_{ref, k}$: current speed at time $k$
\zi \textbf{Input:} $acc_{ref, k}$: expected acceleration at time $k$
\zi \textbf{Input:} $acc_{k}$: actual acceleration at time $k$
\zi \textbf{Input:} \textit{$\Delta_{t}$}: the updated value for $acc_{t}$
\\
\li Fetch original table in the form of $T[cmd][v]=acc$ \label{li:table} 
\li \If Driving Mode != AUTO or 
$\left| \theta_{} \right|$ $ \geqslant\delta_{steer}$
\li \Then break
    \End 
    
\li       Calculate and Preprocess $cmd_{ref, k}, v_{ref, k}, acc_{ref, k}$ \label{li:collect_feedback_data}
\li       Get $acc_{k}$ from sensor
    \End \label{li:oneline_preprocess_done} 
    
\li \If ConvergeCheck == $true$ \label{li:converge1}
\li   \Then break
	\End \label{li:converge2}
\li \For $i \gets 1$ \To $T.Size$ and $j \gets 1$ \To $T[cmd_i].Size$ \label{li:sloop}

\li \Then $acc_{t} \gets$ $T[cmd_{i}][v_{j}]$
\li       Determine $\Delta_{t}$ for each [v][cmd] grid
\li       Updating: $acc_{t}$ = $acc_{t} + \Delta_{t}$ \label{li:updateing}
    \End
\li Export $T'$ from $[cmd][v]=acc$ to $[v][acc]=cmd$ \label{li:updated}
\li Updating: $T \gets T'$ \label{li:new_T}
\li Longitudinal controller $\gets T$ \label{li:apply_table}
\end{codebox}
}
\rule{\linewidth}{1pt}
\end{center}
\end{table}

Data is first preprocessed via Line \ref{li:table} to Line \ref{li:oneline_preprocess_done}. 
If desired speed and actual speed already converges (Eq. \ref{eq:converge_condition}), the algorithm stops calibrating process (Line \ref{li:converge1} to Line \ref{li:converge2}).
\begin{equation}
e_{v}=\left | v_{ref}-v_{actual} \right |\leqslant \gamma_{v}\label{eq:converge_condition}
\end{equation}
where $e_{v}$ is current speed error, $v_{ref}$ is expected speed, $v_{actual}$ is vehicle actual speed and $\gamma_{v}$ is the threshold.

The gain $\Delta_{i}$ (From 
Line \ref{li:sloop} to Line \ref{li:updateing}) is determined by comparing the 
difference between expected acceleration and actual acceleration 
(Eq. \ref{eq:acc_err}).

\begin{equation}
gain_{k}=acc_{ref}-acc_{k} \label{eq:acc_err}
\end{equation}
where $acc_{ref}$ is expected acceleration, $acc_{k}$ is actual acceleration collected from sensors like IMU with delay estimation.
To update calibration table $T$, a cost function described by Eq. \ref{eq:cost_func} is used for each table grid $T[cmd_{i}][v_{j}]$.
\begin{equation}
cost_{cmd_{i},v_{j}} = distance_{cmd_{i}, v_{j}} \times cost_{similarity}
\label{eq:cost_func}
\end{equation}
where $distance_{cmd_{i}, v_{j}}$ refers to distance decay coefficient written as Eq. \ref{eq:distance_func}:
\begin{equation}
\begin{aligned}
distance_{cmd_{i},v_{j}} = \hspace{5.5cm}\\
(1-\mu)\times[\alpha(cmd_{ref}-cmd_{i})^{m_{cmd}} + \beta(v_{ref}-v_{j})^{m_{v}})+\xi]
\end{aligned}
\label{eq:distance_func}
\end{equation}
where $\alpha$ and $\beta$ refer to control command decay coefficient and speed decay coefficient, respectively. 
Both $m_{cmd}$ and $m_{v}$ are distance decay factors.
$\xi$ is set to $1e^{-8}$ 
and $\mu$ is written as Eq. \ref{eq:mu}. 
\begin{equation}
\mu = \begin{cases}
1, & if \left | cmd_{ref}-cmd_{i} \right |\leqslant \delta_{cmd} \text{ }or \left | v_{ref}-v_{j} \right |\leqslant \delta_{v} \\ 
0, & otherwise
\end{cases}\label{eq:mu}
\end{equation}
Apart from distance factor, similarity cost ($cost_{similarity}$) is defined as Eq. \ref{eq:similarity_cost}.
Generally, similarity cost intends to preserve the original table. 
This is because we assume that initial table is reasonably accurate and should 
not be modified significantly during any short time. 
\begin{equation}
cost_{similarity} = \varepsilon \times e^{-\iota \times \left|init\_table[cmd_{i}][v_{j}] - acc_{k}\right|}\label{eq:similarity_cost}
\end{equation}
where $\varepsilon$ is similarity decay coefficient, and $\iota$ is exponential decay factor.
Taken together, the update gain $\Delta_t$ is calculated as 
Eq. \ref{eq:final_update}: 
\begin{equation}
\Delta_{t}=\cfrac{gain_{k} \times \sigma} {1 + cost_{cmd_{i},v_{j}}} \label{eq:final_update}
\end{equation}
where $\sigma$ is learning ratio.

Finally, original Table ($T$) is updated to new table ($T'$) in the form of Eq. \ref{eq:update_constrain} 
(Line \ref{li:updated}), 
\begin{equation}
T'[cmd_{i}][v_{j}] =a_{t}+\Delta_{t} \label{eq:update_constrain}
\end{equation}
subjected to:
\begin{subequations}
\begin{align}
&\cfrac{T'[cmd_{i}][v_{x}]-T'[cmd_{j}][v_{x}]} {cmd_{i}-cmd_{j}} < 0 \text{, }j \text{, }x=1, \cdots, n\\
&\left| cmd_{ref} - cmd_{m} \right| < \delta_{cmd\_gap} \text{, } m = k-t, \cdots, k+t\\
&(v_{ref}-v_{k})\times(a_{ref}-a_{k}) > 0
\end{align}
\end{subequations}
and $T'$ is then ready to be used (Line \ref{li:updateing}). 
As one would expect, $T'$ becomes $T$ in the next control circle 
and a new iteration begins (Line \ref{li:new_T}).

\begin{figure*}[thpb]
\small
\centering
    \subfloat[Throttle]{\label{fig:off_table_throttle}\includegraphics[width= 0.48\linewidth]{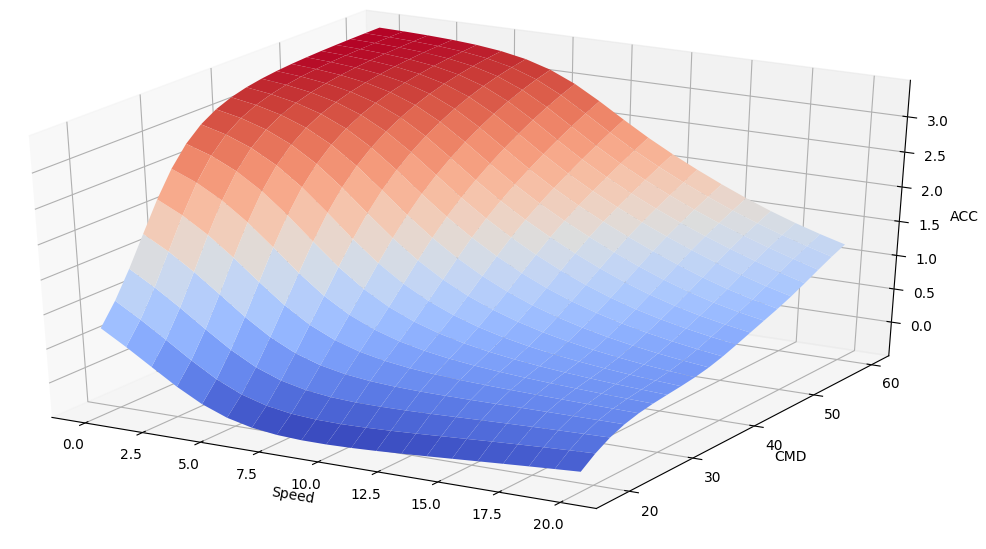}}
    \subfloat[Brake]{\label{fig:off_table_brake}\includegraphics[width= 0.48\linewidth]{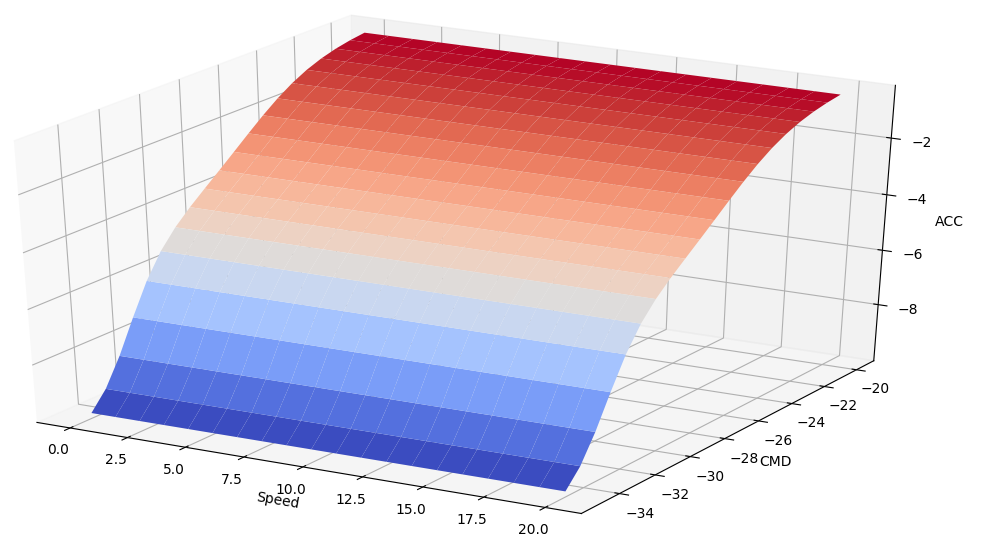}} \\    
  \caption{Offline calibration table 3D view}
  \label{fig:offline_table}
  \scriptsize{}
\end{figure*}

\section{RESULT}\label{sec:result}

\begin{table}[]
\begin{center}
\caption{Vehicle performance of Lincoln MKZ and AX1}
  \label{tab:vehicles}
\begin{tabular}{ | C{0.24\linewidth} | C{0.3\linewidth} | C{0.3\linewidth} |}
\hline
     & Lincoln MKZ   & Neolix AX1 \\ \hline
Power 	& hybrid & electronic\\ \hline
Size($m^3$) & $4.93\times1.86\times1.47$ & $2.68\times1.07\times1.5$ \\ \hline
Weight(kg) & 1,769 & 300 \\ \hline
Load(kg) & N/A & 300 \\ \hline
Type & passenger & cargo \\ \hline
\end{tabular}
\end{center}
\end{table}

\subsection{Test Vehicles}
Two types of vehicles (Table \ref{tab:vehicles}) were used to test the performance under different algorithms:
(1) hybrid passenger vehicle --- Lincoln MKZ;
(2) electronic delivery-only vehicle --- Neolix AX1 \cite{Neolix2018} \footnote{For Neolix AX1, manual driving refers to remote control via a joystick.}.




\begin{table}[]
\begin{center}
\caption{Models Accuracy Comparisons (m/s$^2$)}
  \label{tab:offline_performance}
\begin{tabular}{ | c | c | c | c | c | c | c |}
\hline
& \multicolumn{2}{ |c| }{Throttle} & \multicolumn{2}{ |c| }{Brake} \\ \hline
Model & MAE   & RMSE & MAE & RMSE\\ \hline
Gaussian Process Regression 	& 0.464 & 0.618 & 0.477 & 0.633\\ \hline
Linear Regression 		        & 0.465 & 0.609 & 0.472 & 0.619 \\ \hline
Support Vector Machine	        & 0.465 & 0.615 & 0.471 & 0.621 \\ \hline
M5P					        & 0.230 & 0.313 & 0.230 & 0.312 \\ \hline
Random Forest				& 0.154 & 0.257 & 0.140 & 0.237\\ \hline
\textbf{Neural Network}			& \textbf{0.113}   & \textbf{0.141}  & \textbf{0.141}  & \textbf{0.163} \\ \hline
\end{tabular}
\end{center}
\end{table}

\subsection{Offline Model Evaluation}
Offline model was tested on Lincoln MKZ with
cross validation and road performance.
In this section, test vehicles were put to a medium load, e.g. two passengers plus standard on-board equipments, around 200 kg in total.

Lincoln MKZ has more complex powertrain systems than Neolix AX1. Hence, if the offline calibration model works fine on Lincoln MKZ it should work well on Neolix AX1 too (and it does, see next Section).

\subsubsection{Cross Validation} \label{sec:offline_cv}

Table \ref {tab:offline_performance} shows ten-fold cross-validation results for offline calibration model described in Section \ref {sec:offline_algorithm}. Note that MAE refers to Mean Absolute Error, whereas RMSE refers to Root Mean Square Error.

Among all algorithms tested, Gaussian process regression, linear regression, and support vector machine were first ruled out due to their poor performance.
Random Forest performed considerably better, but best results were obtained using neural network, which, in general, was able to achieve an accuracy within 3\%, compared to $[0, 4]$ $m/s^2$ effective range for throttling and $[-6, 0]$ $m/s^2$ effective range for braking. Taken together, neural network was chosen as the offline calibration model.

\subsubsection{Calibration Table View}

Figure \ref{fig:offline_table} intuitively shows a calibration table modeled by the offline algorithm, where neural network is the modeling algorithm with the same dataset in Section \ref{sec:offline_cv}.
Figure \ref{fig:off_table_throttle} and \ref{fig:off_table_brake} indicate that both the command-acceleration relation and speed-acceleration relation are close to monotonicity, as one may expect.

\subsubsection{Road Performance} \label{sec:offline_hit_the_road}
Usually, algorithms working fine under laboratory conditions with given inputs do not necessary 
mean they would have the same level of performance in actual industrial environments. 
In Baidu, we take road performance very seriously. 
After all, our algorithms will be implemented on numberless vehicles, including but not limited to our own 100+ vehicles.
We used speed error and station error to exam road performance:
(1) speed error  = desired speed - actual speed;
(2) station error = expected location - actual location.


\begin{table}[]
\begin{center}
\caption{Offline Calibration Comparison}
  \label{tab:offline_closeloop_comparision}
\begin{tabular}{|c | c c| c c|}
\hline
         & \multicolumn{2}{c|}{Manual Calibration}     &       \multicolumn{2}{c|}{Auto Offline Calibration}     \\ \cline{2-5}
            & {\tiny Speed MAE ($m/s$)} & {\tiny Station MAE ($m$)} & {\tiny Speed MAE ($m/s$)} & {\tiny Station MAE ($m$)} \\ 
         \hline
\multirow{10}{*}{Rounds} & 0.176       & 0.404         & 0.166       & 0.475         \\
         & 0.133       & 0.523         & 0.168       & 0.371         \\
         & 0.16        & 0.765         & 0.11        & 0.305         \\
         & 0.14        & 0.597         & 0.157       & 0.379         \\
         & 0.15        & 0.615         & 0.162       & 0.399         \\
         & 0.126       & 0.577         & 0.141       & 0.393         \\
         & 0.132       & 0.577         & 0.135       & 0.299         \\
         & 0.151       & 0.65          & 0.151       & 0.319         \\
         & 0.138       & 0.652         & 0.133       & 0.32          \\
         & 0.133       & 0.523         & 0.123       & 0.358         \\  \hline
Average  & 0.144      & 0.588        & 0.145     & 0.362   \\ \hline
\end{tabular}
\end{center}
\end{table}


The offline calibration algorithm was first deployed on a Lincoln MKZ with a standard \href{https://github.com/ApolloAuto/apollo}{Baidu Apollo driving solution}. We then compared road performance between manual calibration\footnote{Manual calibration table was generated using \href{https://github.com/ApolloAuto/apollo/tree/master/modules/tools/calibration}{Apollo calibration tools.}} and offline auto-calibration algorithm. Ten rounds of road tests, each of which consists of 30 minutes with maximum speed at 10 $m/s$, were run for both control group (manual calibration) and experimental group (offline auto-calibration).

The results show that offline auto-calibration model and manual calibration almost performed equivalently on speed error 
(Table \ref{tab:offline_closeloop_comparision}). 
As to station error, however, offline auto-calibration model performed considerably better (Table \ref{tab:offline_closeloop_comparision}).

Table \ref{tab:offline_multi_vehicle} shows more results on randomly selected vehicles (all Lincoln MKZ) for offline auto-calibration model only.
It is clear that the offline model performed similarly across different vehicles, which proves its robustness.

\begin{table}[]
\begin{center}
\caption{Cross-Vehicle Performance}
  \label{tab:offline_multi_vehicle}
\begin{tabular}{|C{0.13\linewidth} | C{0.13\linewidth} C{0.13\linewidth}| C{0.13\linewidth} C{0.13\linewidth}|}
\hline
         & \multicolumn{2}{c|}{Speed ($m/s$)}     &       \multicolumn{2}{c|}{Station ($m$)}     \\ \cline{2-5}
         & MAE & RMSE & MAE & RMSE\\ \hline
\multirow{10}{*}{Vehicle} & 0.140       & 0.159         & 0.313       & 0.286         \\ 
            & 0.158       & 0.177         & 0.354	     & 0.332 \\
            & 0.145       & 0.171         & 0.395       & 0.350 \\ 
            & 0.195       & 0.178        & 0.372	     & 0.325 \\
            & 0.093       & 0.149         & 0.357	     & 0.295 \\
            & 0.170       & 0.175         & 0.339	     & 0.329 \\
            & 0.137       & 0.167         & 0.375	     & 0.335 \\
            & 0.158       & 0.178        & 0.	354       & 0.332 \\
            & 0.150       & 0.182         & 0.379	     & 0.341 \\
            & 0.158       & 0.182         & 0.389	     & 0.339 \\
            \hline
Average  &  0.150     & 0.172        &  0.363    & 0.326   \\ \hline
\end{tabular}
\end{center}
\end{table}


\begin{figure*}[thpb]
\centering
\small
      \subfloat[Speed error without online auto-calibration]{\label{fig:raw_speed_error}
      \includegraphics[width = 0.4\linewidth]{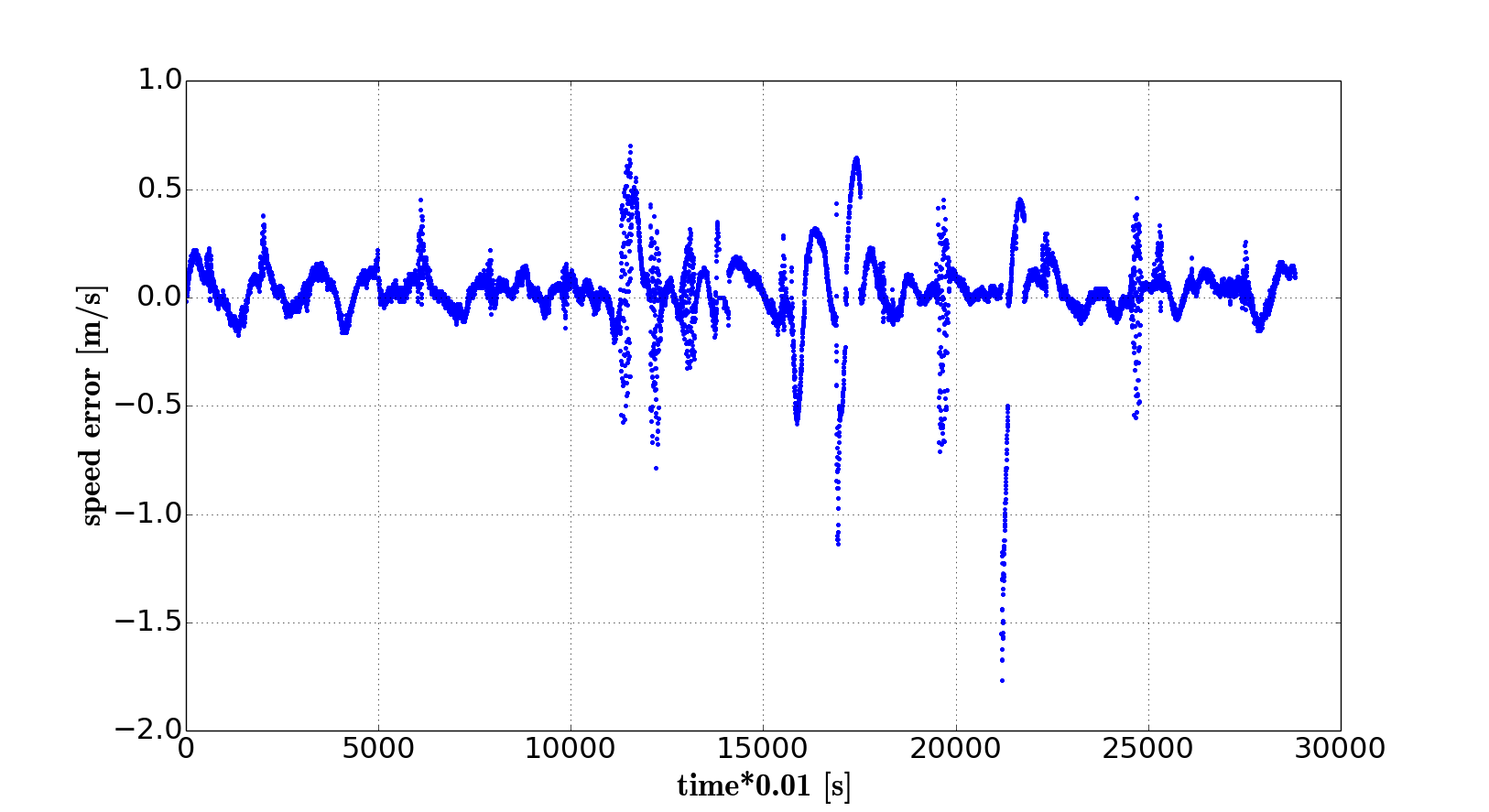}}
      \subfloat[Speed error with online auto-calibration]{\label{fig:online speed_error}
      \includegraphics[width = 0.4\linewidth]{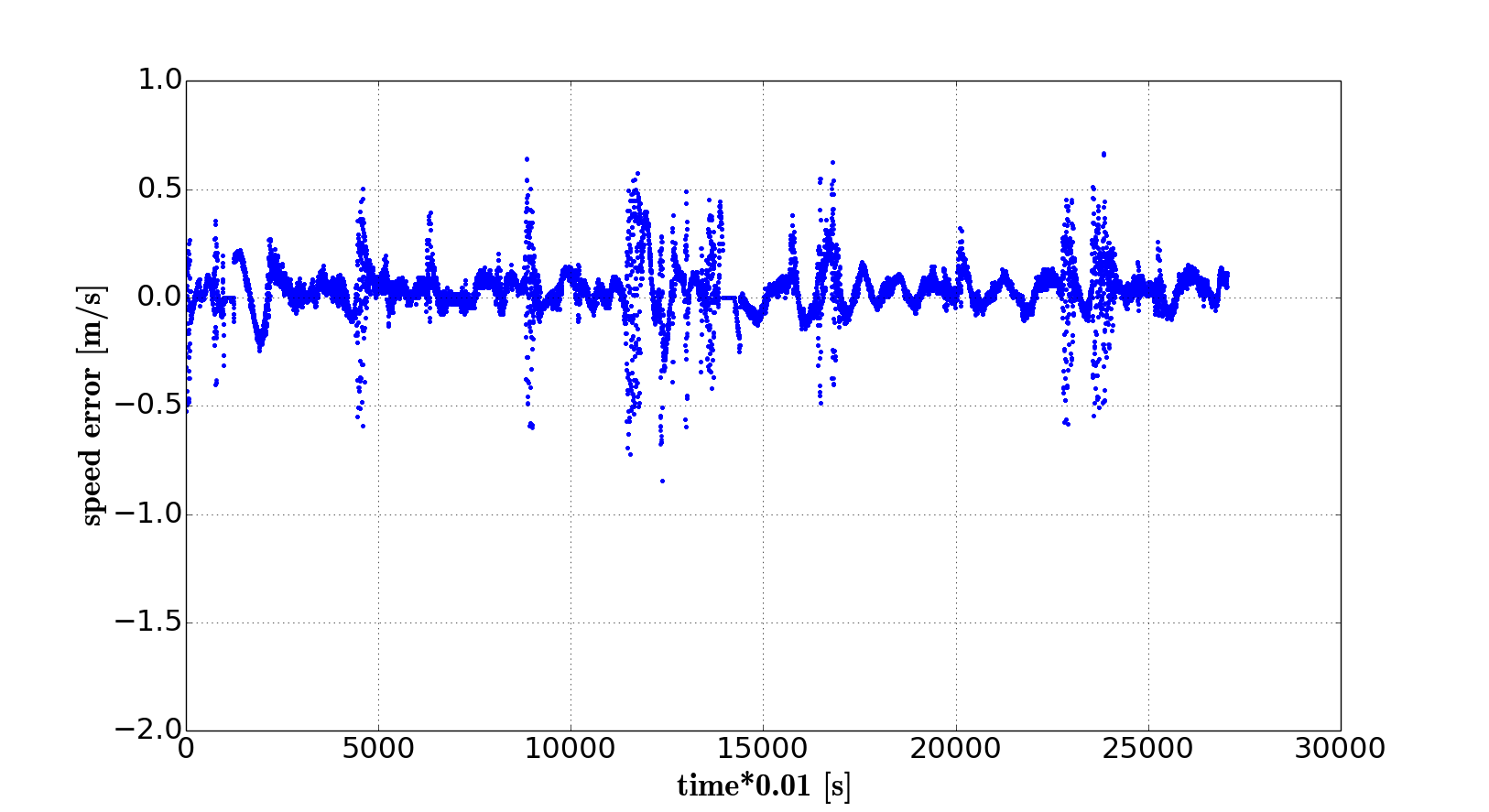}}    \
      \subfloat[Station error without online auto-calibration]{\label{fig:raw_station_error}
      \includegraphics[width = 0.4\linewidth]{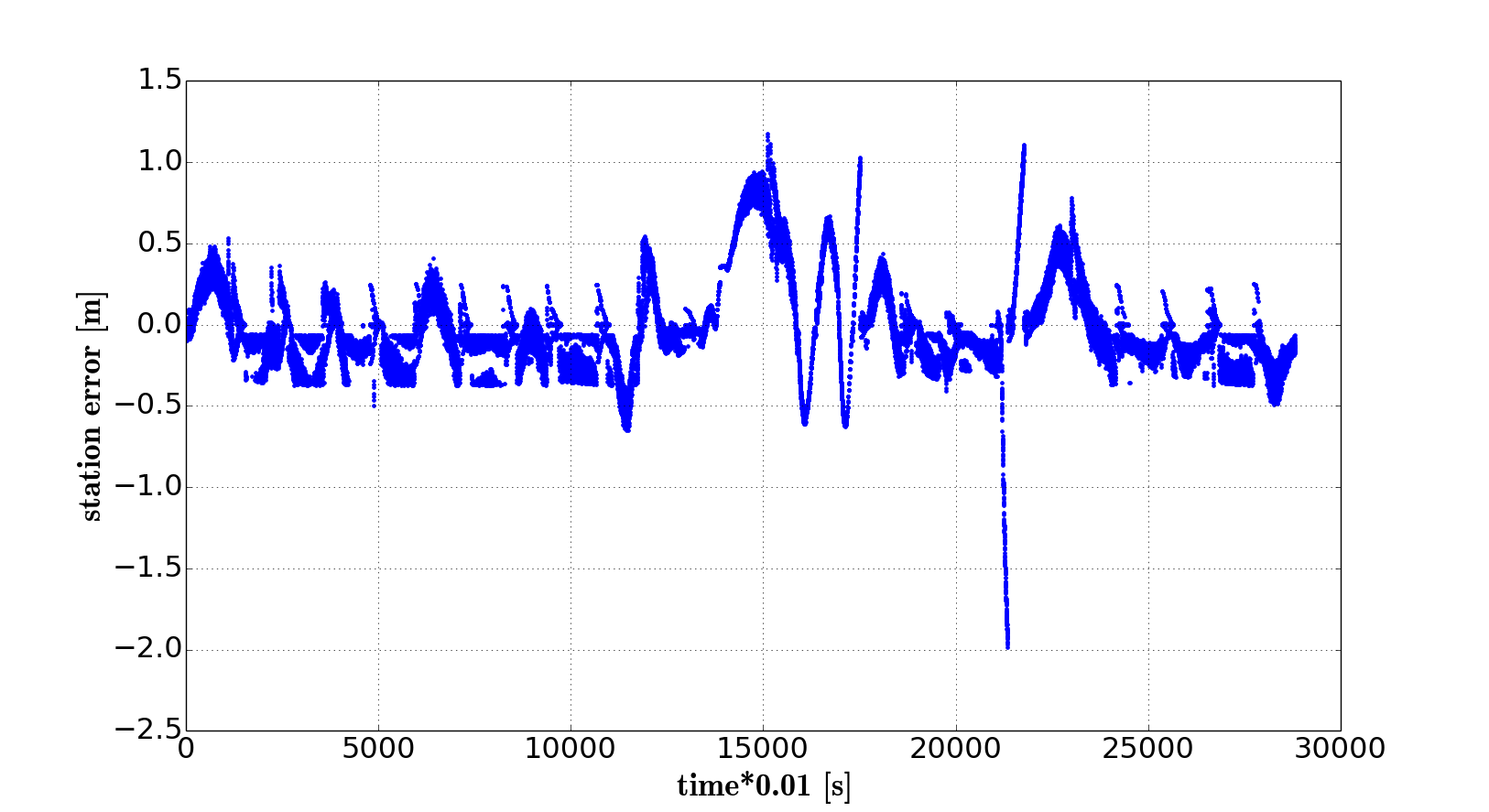}}
      \subfloat[Station error with online auto-calibration]{\label{fig:online_station_error}
      \includegraphics[width = 0.4\linewidth]{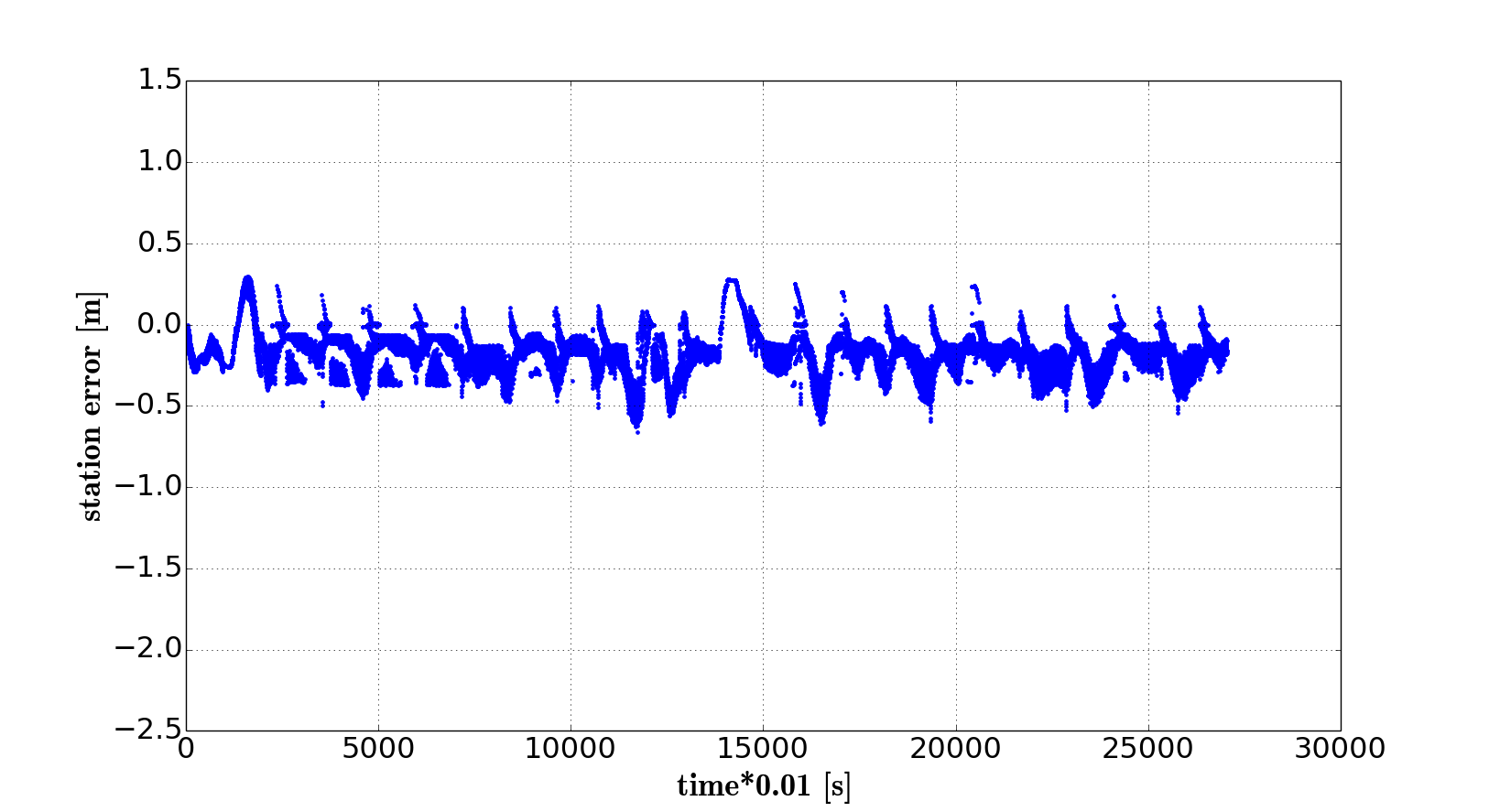}}
      \caption{Control error of AX1 under full load}
      \label{fig:online_error}
\end{figure*}

\subsection{Online Algorithm Evaluation}
For online calibration tests, a Neolix AX1 (a delivery-only vehicle, maximum speed at 3 $m/s$) was used. 
During tests, an offline calibration table was first built at dead load (0 kg), 
then online calibration algorithm started adjusting the table at dead load, 
150 kg load, 300 kg load, and 360 kg load (20\% overload). 
One would expect small speed/station errors across all loads with online calibration enabled.

\begin{table}[thpb!]
\begin{center}
\caption{Speed Errors with Load Changes ($m/s$)}
  \label{tab:load_change_speed}  
\begin{tabular}{ | c | c |  c | c | c |} \hline
& \multicolumn{2}{ |c| }{With online-calibration} & \multicolumn{2}{ |c| }{Without online-calibration} \\ \hline
Load (kg) & MAE   & RMSE & MAE & RMSE\\ \hline
0  & 0.118 &	0.173&   0.126 &	0.184	\\ \hline
150 & 0.114 & 0.165 & 0.105 & 0.152 \\ \hline
300 & 0.102 & 0.153 & 0.149 & 0.199 \\ \hline
360 & 0.113 & 0.157 & 0.154 & 0.206 \\\hline
\end{tabular} 
\end{center}
\end{table}

\begin{table}[thpb!]
\begin{center}
\caption{Station Errors with Load Changes ($m$)}
  \label{tab:load_change_station}  
\begin{tabular}{ | c | c |  c | c | c |} \hline
& \multicolumn{2}{ |c| }{With online-calibration} & \multicolumn{2}{ |c| }{Without online-calibration} \\ \hline
Load (kg) & MAE   & RMSE & MAE & RMSE\\ \hline
0  & 0.214 & 0.245& 0.215 & 0.247	\\ \hline
150 & 0.223 & 0.279 & 0.261 & 0.337 \\ \hline
300 & 0.262 & 0.410 & 0.550 & 0.914 \\ \hline
360 & 0.360 & 0.607 & 0.778 & 1.191 \\\hline
\end{tabular} 
\end{center}
\end{table}

\subsubsection{Road Performance}
Three repeated rounds were set for the test vehicle at each load, with each round consisting of 15 minutes' autonomous driving. 
Results can be found in Table \ref{tab:load_change_speed} and Table \ref{tab:load_change_station}. 
Online calibration outperformed the offline one at almost every load, 
suggesting that online calibration algorithm did adjust the initial table appropriately. Interestingly, improvement on station error again outweighed 
that on speed error, similar to the results of offline calibration tests.
This might due to the fact that commands (throttling/braking) generated from the table only directly affects acceleration. Any error on acceleration will be 
amplified twice to station error (since station error is second integration of acceleration error).



Figure \ref{fig:online_error} shows speed/station error per frame during a typical round at full load (300 kg). Without online calibration, the vehicle's speed fluctuated considerably, with peak value around -2.0 $m/s$. 
Conversely, the vehicle was able to maintain a relatively steady speed with 
online calibration. Similar pattern can also be found on station error.

\subsubsection{Calibration Table View}
A table before and after calibration is available in Appendix in Section \ref{sec:tableview}. 
Note that before calibration the table was simply an initial calibration table 
built from manual driving data at dead load.
Then the vehicle was put at 300 kg load and run for approximately
30 minutes and a new table was generated after thousands of real-time adjustments.

From the updated table, it is clear that same throttling/braking commands lead to
lower acclerations/decelerations. 
This suggests that the vehicle's ability of accleration/deceleration is reduced, 
exactly matching what one would expect when a vehicle is at full load.

%
%
%
%
%
%
%
%
%
%
%
%

\subsection{Computational complexity}
Offline calibration model requires less than ten seconds on a Dell Precision 7510 workstation to build an initial calibration table.
On an industrial standard Nuvo-5095GC machine, online calibration takes less than 2 milliseconds to complete adjustment on calibration table per cycle.
It is also possible to parallelize the computation of grid updating during online calibration, since each grid in a table is independent of other grids \cite{Zhu2012GPGPU}.

\section{CONCLUSION}\label{sec:conclusion}

In this manuscript, we presented a longitudinal control calibration algorithm,  consisting of an offline calibration model and an online calibration algorithm.
For the offline calibration model, we have tested quite a few machine 
learning techniques (see Table \ref{tab:offline_performance}) and eventually found an end-to-end solution (neural network). We have also tried numberless data preprocessing methods to obtain clean and useful data out of massive amount of data and noises (see Section \ref{sec:datapreprocessing}). Contrary to offline calibration, we chose a model-free method for online calibration. Specifically, 
online calibration was implemented in a gradient-descent way (see Section \ref{sec:onlinelearningalg}). 
Since online calibration runs at high frequency (i.e. 100 Hz in our case), gradient-descent 
is a more reliable way to update calibration table, without the risk of choosing a wrong model.

The results show that with calibration enabled, our test vehicles performed 
significantly better as to control accuracy (speed/station error).
What is more, the algorithm has been deployed to multiple Baidu Apollo 
autonomous driving vehicles, including standard hybrid family vehicles and 
electronic delivery-only vehicles. As of August 27th 2018, it had been tested more than two thousands hours, with around ten thousands kilometers' (6,213 miles) road tests.

Finally, we would like to end with Baidu Apollo slogan: \textit{We choose to go to the moon in this decade and do the other things, not because they are easy, but because they are hard. --- John F. Kennedy.}
Baidu Apollo autonomous driving platform is designed to target one of the most challenging problems in the field of artificial intelligence.
We, the Apollo Community, will continue delivering systems that can free drivers' hands.




\renewcommand\thefigure{A.\arabic{figure}}  
\setcounter{figure}{0} 

\section{APPENDIX}\label{sec:appendix}
\subsection{Calibration Table View}\label{sec:tableview}

\begin{figure}[thpb!]

\small
\centering
      \subfloat[initial calibration table]{\label{online calibration raw}
      \includegraphics[width = 0.9\linewidth]{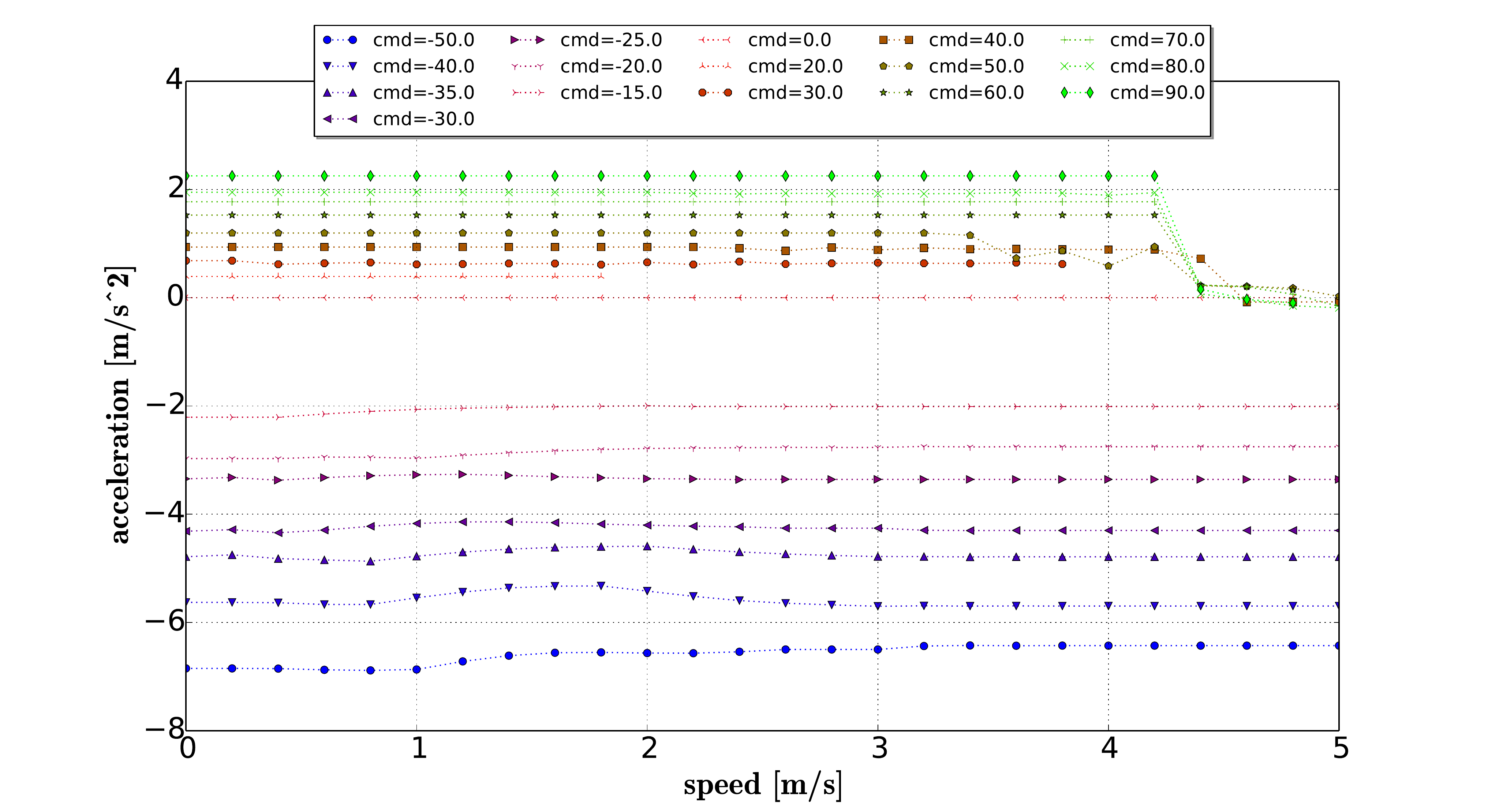}} \

      \subfloat[Online updated calibration table ]{\label{online updated calibration table}
       \includegraphics[width = 0.9\linewidth]{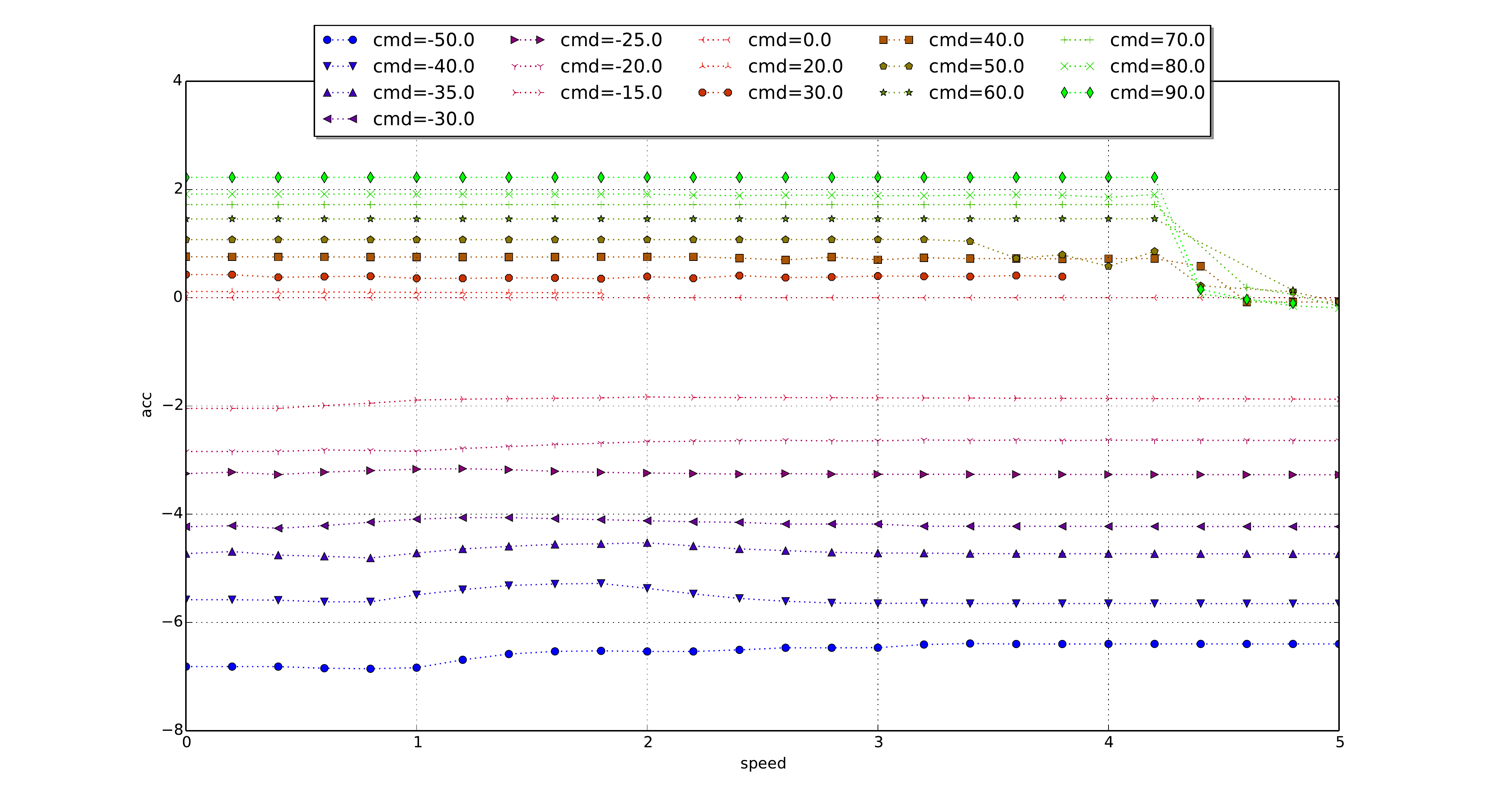}}     
      \caption{Comparison of initial and online updated calibration table}
      \label{fig:table_comparison}
\end{figure}

Figure \ref{fig:table_comparison} illustrates the calibration table before and after the online calibration.
As there is no ground truth for calibration table, whether the updating changes suit our expectation is the criteria here.

Take 20 percent of throttle command as an example, 
it is able to generate an acceleration around 0.5 $m/s^2$ in the initial table, which is inferred without any load.
After online calibrating, system determines that a 20 percent of throttle is only able to generate a slightly bigger than zero acceleration, as one may expected.


\end{document}